\documentclass{article} 
\usepackage{iclr2025_conference,times}
\usepackage{graphicx}
\usepackage[colorlinks=false]{hyperref}
\usepackage{listings}
\usepackage{xcolor}
\usepackage{url}
\usepackage{xurl}
\usepackage{multirow} 
\usepackage{makecell} 
\usepackage{longtable}
\usepackage{array}
\usepackage{tcolorbox} 
\tcbuselibrary{breakable} 

\makeatletter
\def\iclrfinalcopy@footer{} 
\def\iclrfinalcopy@header{} 
\makeatother

\usepackage{amsmath,amsfonts,bm}

\def\eqref#1{equation~\ref{#1}}

\def\1{\bm{1}}

\DeclareMathAlphabet{\mathsfit}{\encodingdefault}{\sfdefault}{m}{sl}
\SetMathAlphabet{\mathsfit}{bold}{\encodingdefault}{\sfdefault}{bx}{n}

\usepackage{hyperref}
\hypersetup{
    colorlinks = false,      
    pdfborder  = {0 0 0},   
    urlcolor   = black,      
}

\lstdefinelanguage{json}{
    basicstyle=\normalfont\ttfamily\small, 
    numbers=left, 
    numberstyle=\scriptsize, 
    stepnumber=1, 
    numbersep=8pt, 
    showstringspaces=false, 
    breaklines=true, 
    breakatwhitespace=true,
    frame=single, 
    rulecolor=\color{black}, 
    backgroundcolor=\color{white}, 
    literate=*{:}{}{0\discretionary{}{}{}}%
             {"}{"}{0\discretionary{}{\break}{}}%
             {/}{/}{0\discretionary{}{}{}}%
             {-}{}{0\discretionary{}{}{}},
    morestring=[b]", 
    stringstyle=\color{string}, 
    upquote=true 
}

\title{MCPToolBench++: A Large Scale AI Agent Model Context Protocol MCP Tool Use Benchmark}

\author{Shiqing Fan, Xichen Ding, Liang Zhang, Linjian Mo \\
Ant Group\\
\texttt{\{zuoran.fsq,xichen.dxc,zhuyue.zl,linyi01\}@antgroup.com} \\
}

\iclrfinalcopy 

\begin{document}

\maketitle
\begin{abstract}

LLMs' capabilities are enhanced by using function calls to integrate various data sources or API results into the context window. Typical tools include search, web crawlers, maps, financial data, file systems, and browser usage, etc. Integrating these data sources or functions requires a standardized method. The Model Context Protocol (MCP) provides a standardized way to supply context to LLMs. However, the evaluation of LLMs and AI Agents' MCP tool use abilities suffer from several issues. First, there's a lack of comprehensive datasets or benchmarks to evaluate various MCP tools. Second, the diverse formats of response from MCP tool call execution further increase the difficulty of evaluation. Additionally, unlike existing tool-use benchmarks with high success rates in functions like programming and math functions, the success rate of real-world MCP tool is not guaranteed and varies across different MCP servers. Furthermore, the LLMs' context window also limits the number of available tools that can be called in a single run, because the textual descriptions of tool and the parameters have long token length for an LLM to process all at once.

To help address the challenges of evaluating LLMs' performance on calling MCP tools, we propose MCPToolBench++\footnote{\url{https://github.com/mcp-tool-bench/MCPToolBenchPP}}\footnote{\url{https://huggingface.co/datasets/MCPToolBench/MCPToolBenchPP}}, a large-scale, multi-domain AI Agent tool use benchmark. As of July 2025, this benchmark is build upon marketplace of over 4k MCP servers from more than 40 categories, collected from the MCP marketplaces and GitHub communities. The datasets consist of both single-step and multi-step tool calls across different categories. We evaluated SOTA LLMs with agentic abilities on this benchmark and reported the results. 

\end{abstract}

\section{Introduction}

\begin{figure}
\includegraphics[height=3.5in, width=5.6in]{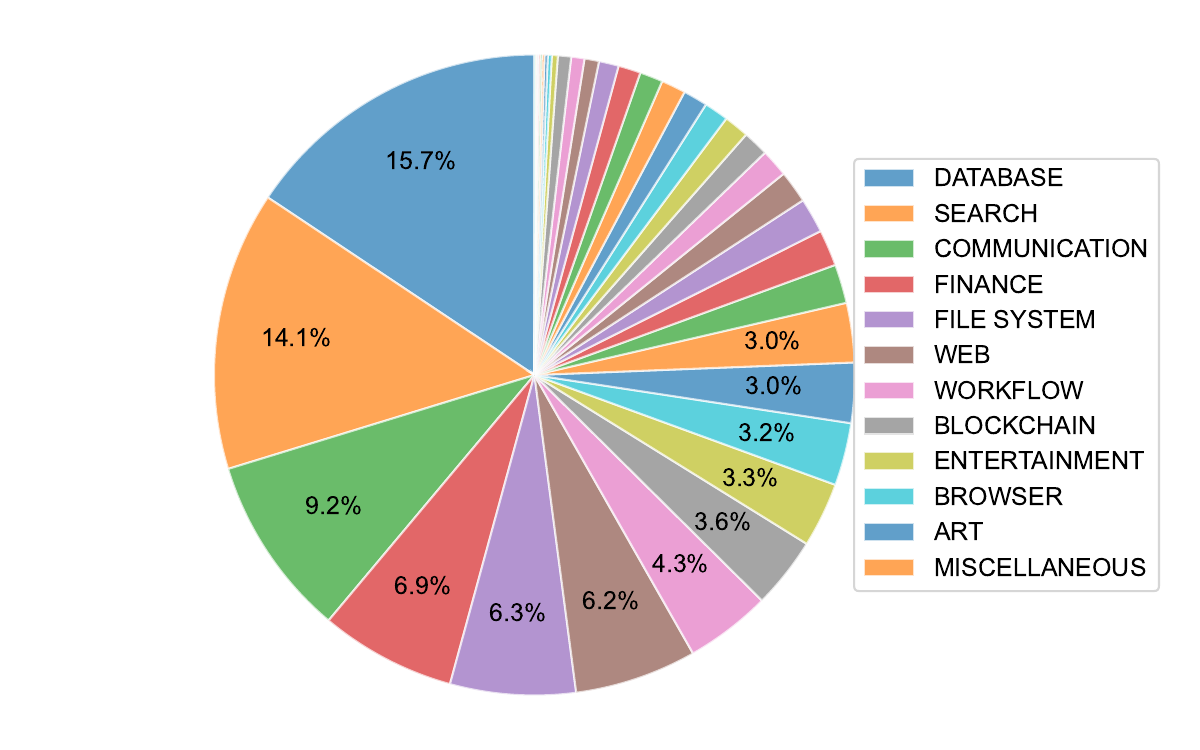}
\caption{Distribution of MCP Servers Count by Category}
\label{fig:category_pie_chart}
\end{figure}

Modern LLMs (\cite{claude4system,openai_o3o4_systemcard,Gemini2.5Report, qwen3}) are transforming from pure language-based models to models with more complicated reasoning and agentic tool use capabilities (\cite{openai25deepresearch,gemini2024deepresearch,claude4,k2-instruct}). Many state-of-the-art LLMs are trained on function-calling capabilities, such as searching, planning, and browse use, to access real-time information (e.g., weather, financial data) and perform complicated tasks like deep research, planning, and API usage (e.g., making reservations). To handle various data sources and function-calling results, \cite{model_context_protocol} provides a standardized way to provide context to LLMs and integrate data sources and tools from various MCP servers. A growing number of MCP servers and clients are available in the marketplaces. However, despite the standardized protocol for LLM context, evaluating the performance of MCP Tool Function Calls remains very difficult due to several aspects:

\textbf{Lack of comprehensive benchmarks and datasets}: Unable to cover highly diverse MCP tools and schemas. Calling MCP tools requires LLMs to select from multiple available tools, which include JSON schemas for tools' descriptions, parameter types, and value schemas. There is a lack of a unified index for data source, tool schemas and server configurations across the marketplace.

\textbf{Parameter Reasoning Capabilities} Many tool descriptions and schemas help LLMs select the best tool and decide appropriate parameter values. However, existing parameters require LLMs to reason about parameter value codes and acronyms, such as stock ticker symbols ($\text{MSFT} \to \text{Microsoft}$, $\text{9988} \to \text{Alibaba in HKEX}$), driving modes (flight/train), and geocodes.

\textbf{Difficulty in evaluating diverse responses} The diverse nature of MCP API responses and user queries goes beyond simple text or images, further increasing the complexity of evaluating the results. For example, when using browser-use MCPs, the $\text{take\_screenshot}$ tool will fulfill a task and save the screenshot image to local storage. Similarly, PayPal's $\text{create\_invoice}$ tool will create an invoice according to user input and send the URL for downloading the invoice to customers. For these task-fulfillment tools, a message indicating the status of the task and related files and supplements are usually provided. 

\textbf{Varied tool success rate and potential risks} Some MCPs registered in communities, particularly those provided by large corporations, offer guaranteed services, while others are less reliable or safe and pose potential risks of prompt attacks and privacy leakage.

To overcome these issues and help scale the evaluation of MCP tool function call abilities, we introduce a new comprehensive benchmark, MCPToolBench++, which consists of 1.5K question-answer pairs covering 6 domains of MCP servers, including search, map, finance, pay, automatic browser usage, file system and more. This benchmark has the following characteristics, making it easy to use and evaluate:

\textbf{Combination of Single-Step and Challenging Multiple-Steps Questions} The dataset combines single-step questions with more complicated and challenging multi-step tool usage questions. Solving multi-step problems requires a sequence of tool call chains, with some utilizing up to 10 tools in a single user request.

\textbf{Diverse and General Agent Capabilities Evaluation} This benchmark dataset evaluates the highly diverse and general capabilities of LLM models and agent systems. These capabilities vary greatly and are invoked through MCP server integration, such as search, planning, browser usage, crawling, API usage, and order creation/processing.

\textbf{Multiple Domains and Multilingual Support} To cover more domains of tasks and scale the size of dataset, the benchmarks are prepared using a pipeline process, which involves selecting MCP tools from over 40 categories in the marketplace as in Figure \ref{fig:category_pie_chart}, cleaning MCP configurations and tool schemas, and retaining high-quality MCP servers and tools. Typical categories include  browser use, file system, finance, pay, etc. Additionally, agents and LLMs should have multilingual ability, making decision on which tool to choose considering the preferred language the question is using automatically. We prepared several multilingual datasets, such as finding routes using maps world-wide, seeking financial data from global financial markets, etc.

\textbf{MCP Tool Call Run Environment and Success Rate Evaluation}
Most official and community MCP servers require specific environments to execute tool calls, such as MCP Clients to initiate the service and agents' workflow to process queries and start the tool calls. We have thoroughly analyzed all tools in the benchmark and verified that those we utilize are tested and offer either free access or sufficient free call quotas from the MCP service provider to enable reproduction of our results on the benchmarks.

\section{MCPToolBench++ Framework}

\begin{figure}
\includegraphics[height=3.84in, width=5in]{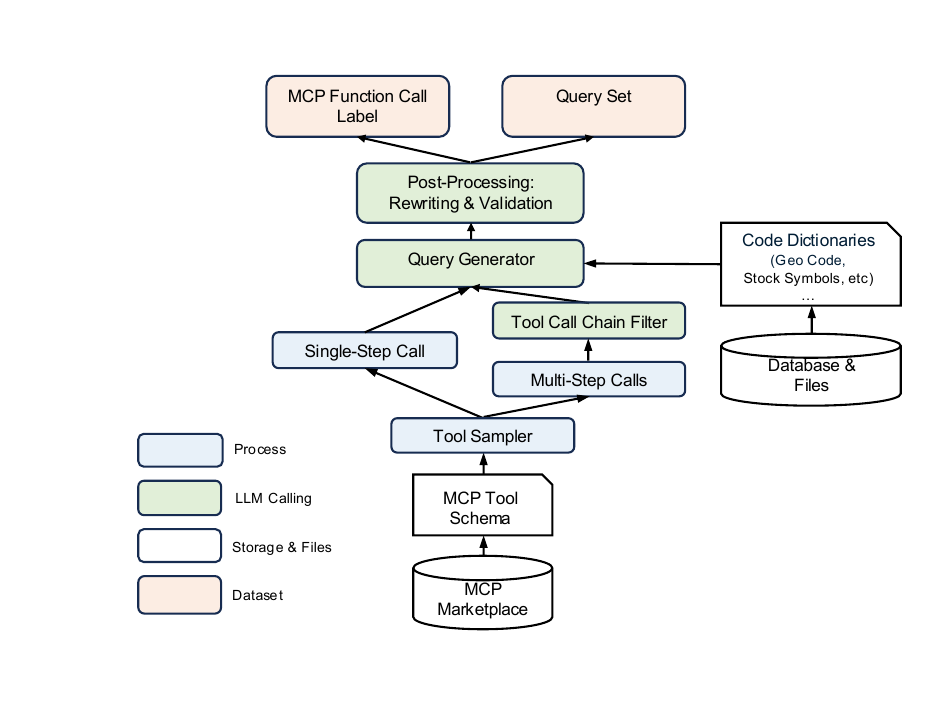}
\caption{Overview of the MCPToolBench++ Data Preparation Process}
\label{fig:main_process_chart}
\end{figure}

\subsection{Overview}

In this section, we will give an overview of the MCPToolBench++ benchmark preparation process in Figure \ref{fig:main_process_chart}. Firstly, we will discuss the data preparation process, including how we collect the MCP servers’ meta-information, MCP configuration files, and tool schemas from various MCP marketplaces. Secondly, based on the MCP tool schema, a tool sampler will sample tools from the schema index, covering both single-step tool calls and multi-step tool calls (or chain of tool calls). For example, query “Get real-time stock prices of Tesla and Microsoft, plot the chart, and calculate todays' change” requires usage of tools $\text{getting\_stock\_data}$, $\text{plot\_chart}$, and $\text{calculator}$. An LLM will generate meaningful prompt templates with slot values ($\text{origin}$, $\text{destination}$, $\text{mode}$), as well as candidates for these slots, e.g., $\text{\{"origin": "JFK Airport", "destination": "Times Square"\}}$. The top-K valid slot values will be selected according to the description of the parameters. Thirdly, a Query Generator will take the sampled tools, the prompt template and parameters' value are generated from the sampled tools' schema. After slot-filling the queries will be further rewritten. Finally, after several reasonability filtering steps in the post-processing, rewrite and validation process, the generated queries are filtered, and a valid query set and the ground-truth labels in JSON format are generated.

\subsection{MCP Servers and Tools Schema Collection}

MCP servers are gathered from the open MCP marketplace across various websites, including smithery.ai\footnote{\url{https://smithery.ai}}, deepnlp.org\footnote{\url{https://www.deepnlp.org/store/ai-agent/mcp-server}}, pulsemcp.com\footnote{\url{https://www.modelscope.cn/mcp}}, and modelscope.cn\footnote{\url{https://www.pulsemcp.com}}. Some of these marketplaces offer direct API support. For schema JSON files, we utilize the mcp-marketplace Python SDK\footnote{\url{https://github.com/AI-Agent-Hub/mcp-marketplace}} to search and filter relevant MCPs by categories such as map and browser-related MCPs. We collected three types of MCP schemas: mcp\_config.json, server\_meta.json, and tool\_schema.json, and indexed them locally for listing and retrieval. For detailed definitions of these MCP server JSON files, please refer to the appendix \ref{file:mcp_config}.

\subsection{Tool Sampler}

After collecting tool schemas from the marketplace, we apply the Tool Sampler to generate single-step tool calls (e.g., sampling the $\text{get\_weather}$ tool) and multi-step tool calls (e.g., $\text{navigate} \to \text{screenshot}$).

Given a total of $N$ tools collected from $M$ MCP servers and $C$ categories, we let $N_{c}$ denote the number of tool schemas collected from the $c$-th category. We apply different sampling strategies for single-step and multi-step tool calls.

For single-step tool calls, we sample tool $t_{i}$ from $N_{c}$ candidates. We use a sampling with replacement strategy, where each tool is sampled with equal probability $p_{i}$.

For multi-step tool calls, which require multiple tool calls to fulfill a user's request, we divide the tool call generation into different bins. Each bin represents a different number of tool calls, ranging from $\text{count} = 2$ to $\text{count} = K$ (we set the maximum count $K$ to 10).

Within the same category $c$, we use a sampling without replacement strategy to obtain a non-duplicated list of sampled tools $[t_{1}, \dots, t_{k}, \dots, t_{K}]$, where $t_{k}$ denotes the $k$-th sampled tool up to $K$. The distribution of queries in different bins is illustrated in the evaluation section.

Cross-categories multi-step tool calls: Sometimes, a user's query requires tool calls with different features from various categories. For example, the prompt "get the stock price from magnificent 7 stocks within last week and plot a time-series chart" requires a financial tool call and a chart plotting tool. We first use a Large Language Model (LLM) to generate meaningful category combinations, such as $[\dots, c_{s}, c_{t}, \dots]$ (e.g., (finance, plot), (search, map), etc.). We then apply the same sampling without replacement strategy used for within-category sampling to the tools from these generated category combinations.

\subsection{Query Generator}

The query generator process contains several consecutive steps: Tool Call Template generation, Parameter Values Generation, Slot Filling, and Query Rewriting. We will use the tool $\text{maps\_directions(origin, destination, mode)}$ as an example to illustrate this process.

\textbf{Tool Call Template generation}: A template is generated from sampled tool list $[t_{1}, \dots, t_{k}, \dots, t_{K}]$. For example, for the single-step tool $\text{maps\_directions}$ from Google Maps, a diverse batch of templates like "Can you help me plan the best route from \{origin\} to \{destination\}?" are generated by an LLM.

\textbf{Parameters Value Generation}: To fill the parameters in the template, valid and meaningful parameters values are generated according to the field description in the tool schema JSON file. For instance, \{"origin":"JFK Airport"\} and \{"destination":"Times Square"\} are Point of Interest (POI) names generated by the inherent knowledge of the LLM. 

\textit{Code Dictionaries} are feed to the context of the LLM to help generate the desired parameters. Some MCP tools require specific codes as input to the functions instead of natural language, such as geo code, stock symbols, acronyms, etc. For example, the weather tool takes input of city and state as geo-code ($ \text{CA} \to \text{California} $), global stock data API requires correct symbols ( $ \text{TSLA} \to \text{Tesla}, \text{700} \to \text{Tencent} , \text{600519} \to \text{Kweichow Moutai} $), etc.

\textbf{Slot Filling}: The generated parameter values are filled into the template to complete a query.

\textbf{Query Rewrite}: The generated queries sometimes have grammatical issues and need to pass through a query rewrite stage to become meaningful queries. For example, a generated query after slot filling might be "Compare today’s stock price change of MSFT and TSLA," where the stock ticker is generated and filled. This would then be rewritten to "Compare today’s stock price change of Microsoft and Tesla."

\subsection{Post Processing \& Validation}

Post-processing and validation steps are crucial for the synthetic dataset. In our MCPToolBench++ benchmark, we applied several post-processing steps to filter out low-quality queries.

\textbf{Semantic check}: Queries that are not properly rewritten will be removed in the semantic check step. For example, "Find good restaurants near 40.7°, 70.3°" would be removed because the tools require raw (longitude, latitude) coordinates as input, and the query rewrite steps were unable to convert the coordinates to a POI name.

\textbf{Reasonableness check}: The reasonableness of the generated queries is checked. Any counterfactual queries will cause the API to fail and need to be removed from the dataset. For example, "How can I travel from New York to Tokyo by train?" is counterfactual because Japan is an island, and there is no way to travel from New York to Tokyo by train.

\section{Dataset Diversity \& Analysis}

\subsection{Tool Schema Complexity Analysis}

\begin{table}
  \caption{Statistics of Dataset Instance MCP Tools Count and Tokens}
  \label{tab:mcp_tool_token_statistics}
  \begin{tabular}{|c|c|c|c|c|}
        \hline
        Category & Number Instance & MCP Tool Count & Tokens Per Tool & Total Tokens \\
        \hline
        Browser & 187 & 32 & 107.4 & 3.4 K \\
        \hline
        File System & 241 & 11 & 143.8 & 1.6 K \\
        \hline
        Search & 181 & 5 & 555.6 & 2.8 K \\
        \hline
        Map & 500 & 32 & 401.3 & 13 K \\
        \hline
        Finance & 90 & 1 & 505.0 & 0.5 K \\
        \hline
        Pay & 310 & 6 & 656.5 & 3.9 K \\
        \hline
        Total & 1509 & 87 & 288.3 & 25 K \\
        \hline
    \end{tabular}
\end{table}

We conducted complexity analysis of the MCP function calls that LLM needs to process. 

In the formulation, ${M}$ denotes number of MCP servers installed on AI Agent systems, ${N_{t}}$ denotes the average number of tools that the server provides, and $T_{tool}$ denotes the average token length of the schema description of the tool and parameters. The tokens that each function call LLM needs to process have complexity $\mathcal{O}(MN_{t}T_{tool})$. And $T_{tools}$ usually has magnitude ranging from $0.1K$ to $1K$. The total number of available tools has the magnitude of a few hundreds before ranking or relevance filtering. The statistics of some MCP servers from various domains are listed in Table \ref{tab:mcp_tool_token_statistics}.

\textbf{Tool Dispatcher Retrieval and Relevant Tools Selection}
Tools Dispatcher is needed to retrieve relevant tools from the tool schema description to avoid the explosion of unnecessary tools to LLMs. Tool Dispatcher aims to select relevant tools given users’ original query, which can help reduce the average number of tools consumed by LLM from ${N_{t}}(\sim 100)$ to ${N_{k}} (\sim 10)$, usually from a few hundreds to the magnitude of 10. The overall complexity becomes $\mathcal{O}(MN_{k}T_{tool})$, which can remove unnecessary tools schema and help increase the overall accuracy performance.


\section{Evaluation}

We evaluated the MCPs’ function calling abilities across a wide range of models on the MCPToolBench++ benchmark, including models from the OpenAI \cite{openai2024gpt4technicalreport}, Claude \cite{claude4}, Qwen \cite{qwen3}, Kimi \cite{k2-instruct} and others. The benchmark datasets are further characterized by multilingual support (English, Chinese, French, Russion, etc), different steps of tool calls (single-step vs. multi-steps), and various categories (search, browse, map, etc.) to reflect performance across different levels of task difficulty.

\subsection{Metrics}

\subsubsection{Accuracy Evaluation}

\textbf{Abstract Syntax Tree AST}
To evaluate the accuracy of MCP tool calls by LLMs, we adopted the Abstract Syntax Tree (AST) metric. This metric compares the predicted output with ground truth labels for function match, required parameters match, parameter type and value match (\cite{patil2025bfcl}).

\textbf{Multi-Step AST DAG Accuracy}:  For multi-step tool executions, we apply a modified metric of AST, and proposed the metric \textbf{AST DAG Accuracy}. The metric AST DAG-Accuracy denotes the Directed Acyclic Graph (DAG) evaluation of AST score between the predicted multi-steps tool call execution plan with the ground-truth tool call execution plan. This metric is useful because tool call executions have dependencies on previous tool call results. Some prompts have the DAG structure, for example in prompt "Find the top 10 stocks with highest price increase from market of NASDAQ, NYSE and DOW and London Stock Exchange, and then plot the stock and in a chart.". The execution plans looks like: $\textit{\text{get\_top\_change\_stocks("NASDAQ")}} \to \textit{\text{plot\_chart}}$, $\textit{\text{get\_top\_change\_stocks("NYSE")}} \to \textit{\text{plot\_chart}}$, $\textit{\text{get\_top\_change\_stocks("DOW")}} \to \textit{\text{plot\_chart}}$, $\textit{\text{get\_top\_change\_stocks("LSE")}} \to \textit{\text{plot\_chart}}$. The $\textit{get\_top\_change\_stocks}$ tool runs in parallel and the $\textit{plot\_chart}$ tool depends on the previous 4 tool call results. Note that for AST DAG Accuracy metric, $label=1$ denote the children nodes (last execution nodes of each task) should match between the prediction and the ground-truth, and $label = 0$ denotes otherwise. We also notice the phenomena that different models tend to choose different paths of tool calls to finish the same tasks. For example the route planning task, some LLMs tend to choose one step of direct tool call that takes in address as input to the tool \textit{ plan\_route(origin\_address, destination\_address)}, and other LLMs tend to divide the task into three steps, converting the origin and destination address to geocodes or coordinates first and then call the \textit{plan\_route(origin\_geocodes, destination\_geocodes)}.

$$\text{AST DAG Accuracy} = \text{calculate\_dag\_accuracy}(\text{DAG}_{predict}, \text{DAG}_{ground\_truth})$$

\subsubsection{Tool Call Evaluation}

\textbf{$\text{Pass@K}$ Accuracy}: To measure whether the execution result from the MCP tool call is correctly aligned with the expected output, we use $\text{Pass@K}$ to evaluate the accuracy of the MCP function call. A successful MCP tool call run has many conditions, including the input parameters should be correct, the tool call should be successful, the execution results should not be empty and align with the expected ground-truth, etc. To evaluate whether the MCP tool call (API calling) succeeded or failed with errors, we evaluate the response status of each MCP tool call. Sometimes, the MCP server return a 200 success status code, but the detailed message is showing an internal parameter error, such as \{"success": false, "data": null, "error": "[{'type': 'text', 'text': 'Geocoding failed... '}]"\}. To evaluate the detailed message and whether the expected output is aligned, we use an LLM as judge to label the success status of each response. Some root-cause analysis of errors and related prompts are provided in the appendix \ref{tab:mcp_failure_root_causes}.

\begin{figure}
\includegraphics[height=3.42in, width=6in]{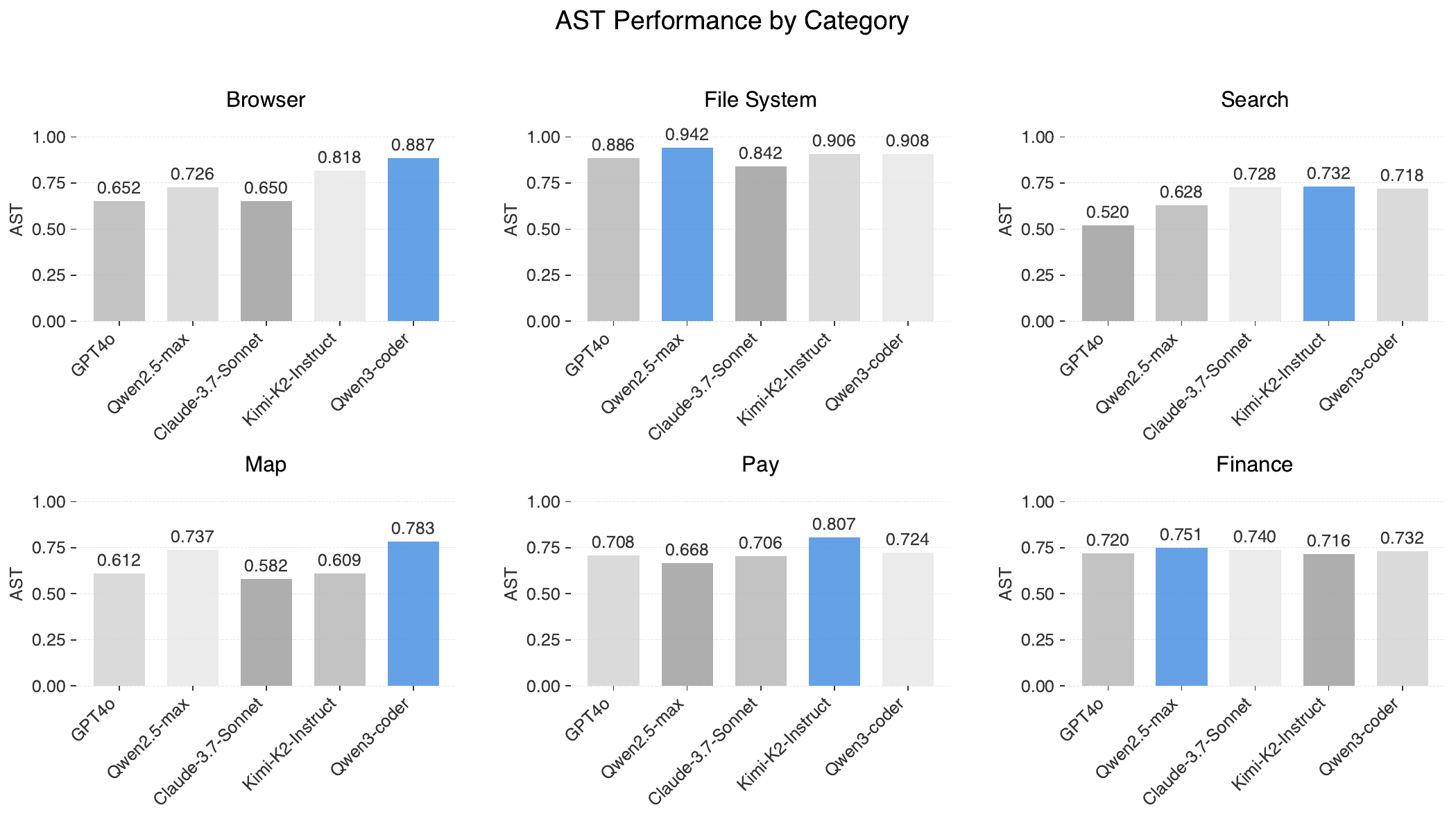}
\caption{MCP Tool Call Abstract Syntax Tree Score Performance}
\label{fig:mcp_tool_call_ast_score}
\end{figure}

\begin{figure}
\includegraphics[height=3.42in, width=6in]{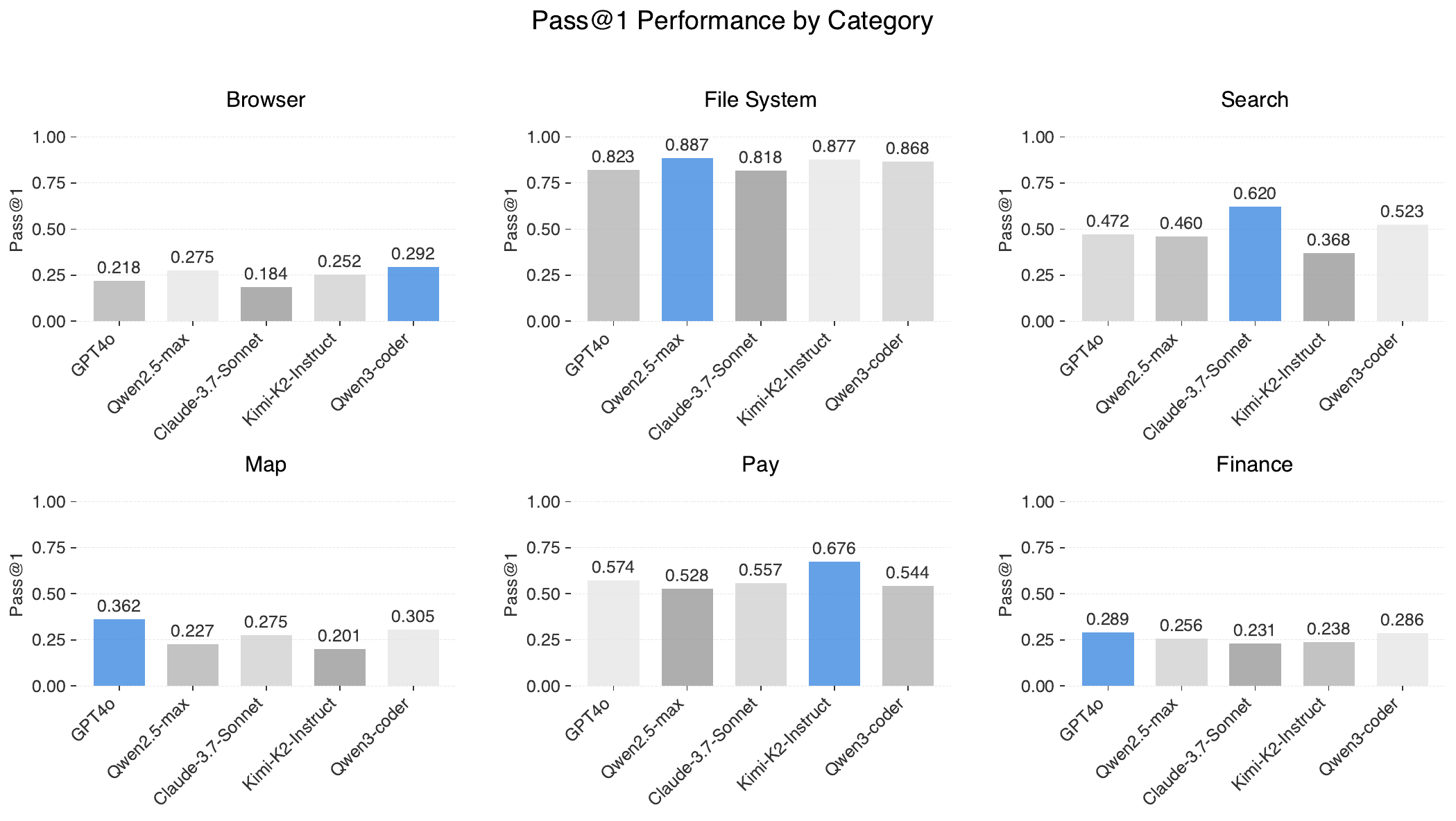}
\caption{MCP Tool Call Pass@K Score Performance}
\label{fig:mcp_tool_call_pass_k_score}
\end{figure}

\begin{table}[htbp]

\caption{Evaluation of MCP Tool Call Abstract Syntax Tree (AST) and Pass@K}
\centering
\begin{tabular}{|l|r|r|r|r|r|r|r|r|}
\hline 
& \multicolumn{2}{c|}{Browser} & \multicolumn{2}{c|}{File System} & \multicolumn{2}{c|}{Search} \\
\hline
Model & AST & Pass@1 & AST & Pass@1 & AST & Pass@1  \\
\hline
GPT-4o & 0.6524 & 0.2182 & 0.8863 & 0.8232 & 0.5200 & 0.4720 \\      
\hline
Qwen2.5-max & 0.7262 & 0.2749 & 0.9419 & 0.8871 & 0.6280 & 0.4600 \\
\hline
Claude-3.7-Sonnet & 0.6503 & 0.1840 & 0.8415 & 0.8183 & 0.7280 & 0.6200 \\
\hline
Kimi-K2-Instruct & 0.8182 & 0.2524 & 0.9062 & 0.8772 & 0.7320 & 0.3680 \\
\hline
Qwen3-coder & 0.8866 & 0.2925 & 0.9080 & 0.8680 & 0.7180 & 0.5227 \\
\hline
& \multicolumn{2}{c|}{Map} & \multicolumn{2}{c|}{Pay} & \multicolumn{2}{c|}{Finance} \\
\hline
Model & AST & Pass@1 & AST & Pass@1 & AST & Pass@1  \\
\hline
GPT-4o & 0.6120 & 0.3616 & 0.7077 & 0.5742 & 0.7200 & 0.2889  \\
\hline
Qwen2.5-max & 0.7372 & 0.2272 & 0.6684 & 0.5277 & 0.7511 & 0.2556 \\
\hline
Claude-3.7-Sonnet & 0.5820 & 0.2748 & 0.7058 & 0.5574 & 0.7400 & 0.2311 \\
\hline
Kimi-K2-Instruct & 0.6088 & 0.2008 & 0.8071 & 0.6761 & 0.7156 & 0.2378 \\
\hline
Qwen3-coder & 0.7830 & 0.3054 & 0.7240 & 0.5440 & 0.7320 & 0.2860
 \\
\hline
\end{tabular}
\label{tab:evaluation_ast_pass_k}
\end{table}

\begin{figure}
\includegraphics[height=3.9285in, width=5.5in]{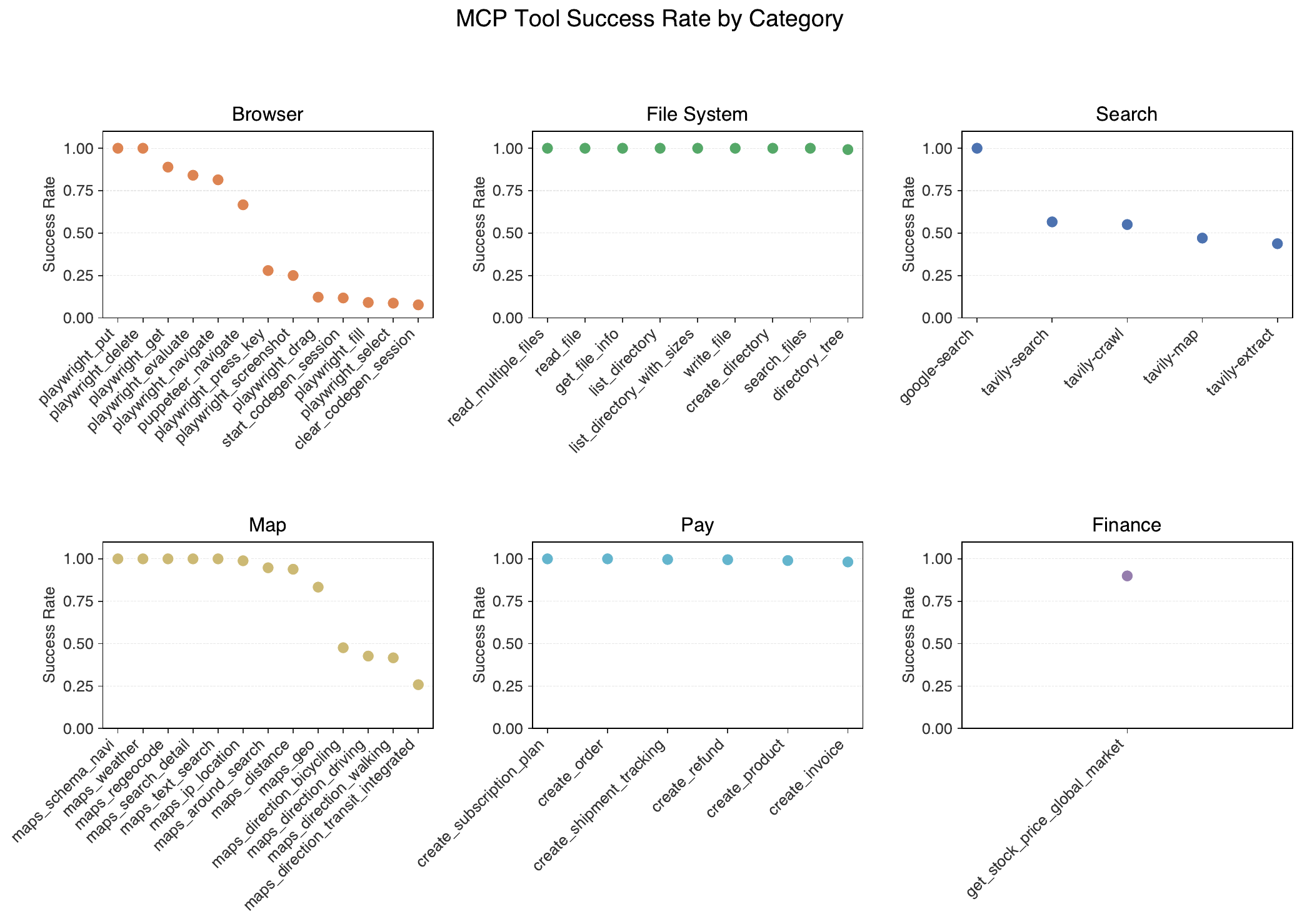}
\caption{MCP Tool Run Success Rate By Category}
\label{fig:mcp_tool_call_success_rate}
\end{figure}

\textbf{Tool Call Success Rate}
The tool call success rate is a dynamic metric that measures the ratio of successful tool call runs to total runs for each MCP tool. In the MCP tool call scenario, this metric is defined as whether the tool call is successfully executed, which is measured by several conditions including whether the response status code contains success status codes (e.g., 200) and whether the "success" key of the result dictionary in the MCP JSON RPC response is "true", etc. It is important to note that the Tool Call Success Rate and Pass@K are closely related metrics. The key difference between them lies in the following: Tool Call Success Rate only measures whether the tool call run is successful without any errors, whereas Pass@K not only measures the success of the tool call run but also the correctness of input parameters and whether the returned results (both the result data and system messages) align with the expected ground truths.

\subsection{Results}

We reported the MCP tool call Abstract Syntax Tree (AST) and Pass@K scores on the evaluated MCPToolBench++ benchmark in Table \ref{tab:evaluation_ast_pass_k}. We illustrate the AST results in Figure \ref{fig:mcp_tool_call_ast_score}, Pass@K results in Figure \ref{fig:mcp_tool_call_pass_k_score} and Tool Call Success Rate in Figure \ref{fig:mcp_tool_call_success_rate}. From the result of Abstract Syntax Tree score (function call tool choosing and parameters inferences), we can see that Qwen3-coder achieved top AST accuracy performance in categories including Browser and Map. And Qwen2.5-max achieved top AST accuracy performance in categories including File System and Finance. And Kimi-K2-Instruct achieved top AST accuracy performance in categories of Search and Pay. To analyze the Pass@1 tool call execution performance, Qwen3-coder achieved top Pass@1 score in the Browser category. Qwen2.5-max achieved top Pass@1 score in the File System category. Claude-3.7-Sonnet achieved top Pass@1 in Search category. GPT-4o achieved top Pass@1 in Map and Finance categories. And Kimi-K2-Instruct achieved the top Pass@1 in the Pay category.

\subsubsection{AST Score vs. Pass@K}

The AST score evaluates how well models choose among tools and make function calls by filling parameters according to the tool schema and descriptions. After the LLM's function call, Pass@1 further evaluates how well we actually run the MCP tool using the parameters inferred by the model. We notice that the \textbf{Tool Call Success Rate} of MCP tool is one of the key variables influencing the overall Pass@1 score, especially when the tools require API connections. To get more accurate Pass@K score, we repeat the experiments multiple times and run multiple trials per each tool call. In the experiments, we set the hyper-parameter of trials per tool call as 5.

Furthermore, we notice that the rankings of AST scores and Pass@K scores are not always positively correlated. For example, in the Search category, Claude-3.7-Sonnet achieved second place in the AST score (0.728) after Kimi-K2-Instruct (0.732). However, after actual tool call runs, Claude-3.7-Sonnet achieved first place in Pass@1 (0.620) compared to Kimi-K2-Instruct (0.368). The results showed that the Google Custom Search Tool (google-search) has a higher success rate than other MCP search tool providers such as Tavily, etc. Additionally, Claude-3.7-Sonnet selects google-search more frequently than other tools compared to other models. This highlights situations, when multiple tools with similar features (web-search, web-crawling, location-based map services) are available, agent models may choose either of the tools and match the ground-truth labels for abstract syntax tree scores, but have large gaps in final results after actually running the tools.

\begin{figure}
\includegraphics[height=2.5in, width=5in]{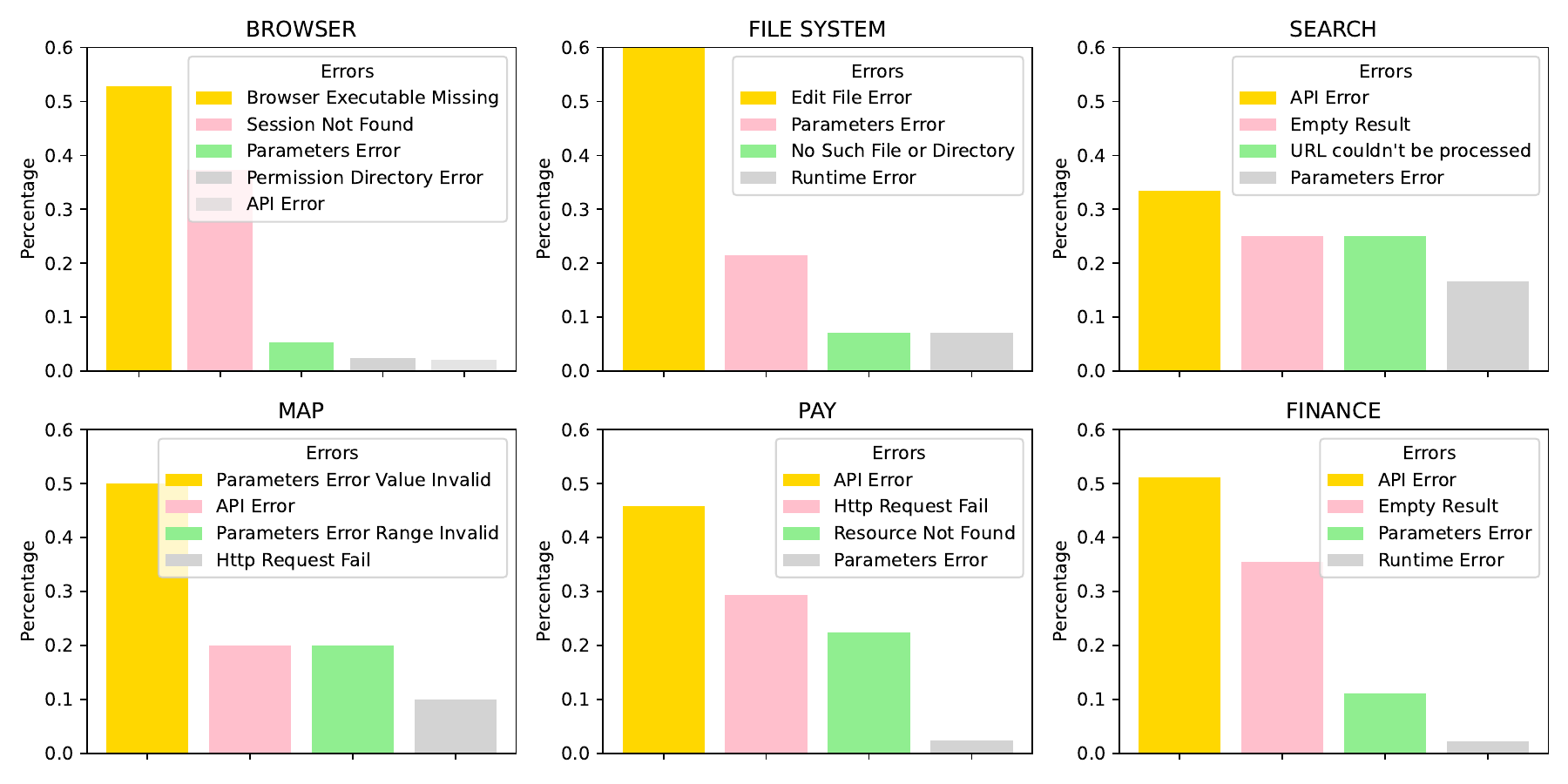}
\caption{MCP Tool Call Errors Root Cause Analysis}
\label{fig:error_analysis}
\end{figure}

\subsection{ERROR ROOT CAUSE ANALYSIS}

We have done a detailed analysis of MCP tool call execution logs and categorize the error codes and messages for each domain. The top ranked root causes of MCP tool call failures are summarized in Figure \ref{fig:error_analysis}, including ”Parameter Errors”, ”API Error”, ”Empty Result”, ”Session \& Runtime Errors”, etc. Also, each domain also has some specific errors. For example, the MCPs in ”MAP” domain have errors "parameters errors range invalid", indicating the longitude and latitude values inferred by LLM function calls are out of range, MCPs in ”Browser Use” domain have errors of ”No Such File or Directory” when taking screenshots of webpage and saving to non-exist local filepaths, etc. MCPs in ”Search” domain have URL processing errors from fetching and extracting the invalid URL or URL needs second-time authentification of web contents, etc. We also list more detailed examples of root causes of MCP tool call errors in the Appendix \ref{tab:mcp_failure_root_causes}.

\section{Related Work}

\subsection{Tool Use Agent}
Tool use agents have accelerated the productivity and increase performance of many real-world tasks. Coding agents\footnote{\url{https://www.deepnlp.org/store/ai-agent/coding-agent}} (\cite{geminicle, openai_codex, claude_code}) apply agentic function calls to manage, edit, generate codes using command lines to fulfill tasks. Deep research agents (\cite{gemini_deep_research,openai_deep_research}) utilize multiple steps of tool uses including searching and crawling to generate detailed analysis of users' complex questions. Moreover, with the development of standardized model context protocol, more tools such as search, browser use, location services can be integrated to LLM based systems.

\subsection{Function Call and MCP Benchmarks}

Function call benchmarks are crucial in evaluating tool use and function call capabilities of LLMs and AI agent systems \cite{mialon2023gaia,patil2025bfcl}. GAIA is a benchmark to test Agent tool use abilities of prompting and search \cite{mialon2023gaia}. Berkeley Function Calling Leaderboard (BFCL) \cite{patil2025bfcl} is a benchmark to test functions call of various programming languages, and uses Abstract Syntax Tree (AST) metric to evaluate the static accuracy of serial and parallel functions. ComplexFuncBench \cite{zhong2025complexfuncbenchexploringmultistepconstrained} focus on the multi-step user constrained function calling under long-context scenario. Some other research focus on generating high quality datasets using pipelines from API repos (\cite{liu2024apigenautomatedpipelinegenerating}), and testing various capabilities such as the browser use (\cite{wei2025browsecompsimplechallengingbenchmark}). 

With the adoption of model-context protocol, more MCP benchmarks are proposed (\cite{luo2025evaluationreportmcpservers,gao2025mcpradarmultidimensionalbenchmarkevaluating}) on different domain such as math, coding, etc.

\section{Conclusion}
In this paper, we introduce MCPToolBench++ benchmark to evaluate LLM and AI Agents' abilities to use MCP tools correctly. The benchmark covers wide range categories of real world tools, which provide multi-domains and multilingual supports. To scale the MCP tool benchmark preparation, we introduce an automatic pipeline build upon mcp configs and tool schemas collected from open MCP marketplaces. Additionally, multiple SOTA models are evaluated on the benchmarks. And detailed success rate and root cause analysis of MCP tool call errors are conducted to highlight future improvements.

\bibliography{iclr2025_conference}
\bibliographystyle{iclr2025_conference}

\appendix

\section{Appendix}

\subsection{Examples of MCP Configs and Schemas}
In this section, we will use $\text{google-maps}$ MCP to show you the format of the schema of servers, tools and configurations. 

\label{file:mcp_config}
\begin{lstlisting}[]
{
  "mcpServers": {
    "google-maps": {
      "command": "npx",
      "args": ["-y", "@modelcontextprotocol/server-google-maps"],
      "env": {
        "Maps\_API\_KEY": "<YOUR\_API\_KEY>"
      }
    }
  }
}
\end{lstlisting}



\begin{lstlisting}[]
{
    "id": "modelcontextprotocol/servers/google-maps",
    "content_name": "Google Maps",
    "category": "MAP",
    "description": "The Google Maps MCP Server offers various tools for interacting with the Google Maps API, including geocoding, place searching, distance calculations, elevation data, and directions, requiring a Google Maps API key for setup and usage with platforms like Claude Desktop or VS Code.",
    "website": "https://github.com/modelcontextprotocol/servers-archived",
    "content": " .. README.md ...",
    "github": "https://github.com/modelcontextprotocol/servers-archived"
}
\end{lstlisting}


\begin{lstlisting}[]
{
    "tools": [{
        "name": "maps_directions",
        "description": "Get directions between two points",
        "input_schema": {
            "type": "object",
            "properties": {
                "origin": {
                    "type": "string",
                    "description": "Starting point address or coordinates"
                },
                "destination": {
                    "type": "string",
                    "description": "Ending point address or coordinates"
                },
                "mode": {
                    "type": "string",
                    "description": "Travel mode (driving, walking, bicycling, transit)",
                    "enum": [
                        "driving",
                        "walking",
                        "bicycling",
                        "transit"
                    ]
                }
            },
            "required": [
                "origin",
                "destination"
            ]
        }
    }]
}
\end{lstlisting}

\subsection{MCP Servers and Tools By Categories}

In this section, we will list some of the top selected MCP servers and their tools by categories in the MCPToolBench++ benchmark.

\begin{longtable}{|c|p{5cm}|p{6cm}|}

  \caption{MCP Servers By Various Categories}
  \label{tab:mcp_servers_by_category} \\
  \hline
  \textbf{Category} & \textbf{Server} & \textbf{Tools} \\
  \hline
  \endfirsthead
  
  \multicolumn{3}{c}{{\bfseries Continue \thetable} } \\
  \hline
  \textbf{Category} & \textbf{Server} & \textbf{Tools} \\
  \hline
  \endhead
  
  \hline
  \endfoot
  
  \hline
  \endlastfoot

  \multirow[t]{2}{*}{\shortstack{BROWSER}}
  & puppeteer/puppeteer & 
  \shortstack[l]{
    puppeteer\_navigate \\ 
    puppeteer\_screenshot \\ 
    puppeteer\_click \\ 
    puppeteer\_fill \\ 
    puppeteer\_select \\ 
    puppeteer\_hover \\ 
    puppeteer\_evaluate} \\
  \cline{2-3}
  
  & executeautomation/playwright & 
  \shortstack[l]{
    playwright\_navigate \\
    playwright\_screenshot \\
    playwright\_click \\
    playwright\_iframe\_click \\
    playwright\_iframe\_fill \\
    playwright\_fill \\
    playwright\_select \\
    playwright\_hover \\
    playwright\_get \\
    playwright\_post \\
    playwright\_put \\
    playwright\_press\_key \\
    playwright\_save\_as\_pdf \\
    playwright\_click\_and\_switch\_tab} \\
  \hline

  \multirow[t]{2}{*}{\shortstack{FILE SYSTEM}}
  & modelcontextprotocol/filesystem & 
  \shortstack[l]{
    read\_file  \\
    read\_multiple\_files  \\
    write\_file  \\
    edit\_file  \\
    create\_directory  \\
    list\_directory \\
    list\_directory\_with\_sizes \\
    directory\_tree \\
    search\_files \\
    get\_file\_info \\
    list\_allowed\_directories} \\
    \cline{2-3}

  \hline

  \multirow[t]{2}{*}{\shortstack{SEARCH}} 
  & adenot/mcp-google-search & 
  \shortstack[l]{
    search \\ 
    read\_webpage} \\
    \cline{2-3}
  
  & tavily/mcp\_tavily-mcp & 
  \shortstack[l]{
    tavily-search \\ 
    tavily-extract \\ 
    tavily-crawl \\ 
    tavily-map} \\
  \cline{2-3}
  
  & mendableai/firecrawl-mcp-server & 
  \shortstack[l]{
    firecrawl\_search \\ 
    firecrawl\_scrape \\ 
    firecrawl\_map \\
    firecrawl\_crawl \\
    firecrawl\_check\_crawl\_status \\
    firecrawl\_extract \\
    firecrawl\_deep\_research \\
    firecrawl\_generate\_llmstxt} \\
  \hline

  \multirow[t]{3}{*}{\shortstack{MAP}}
  & google-maps/google-maps & 
  \shortstack[l]{
    maps\_directions \\
    maps\_geocode \\
    maps\_reverse\_geocode \\
    maps\_search\_places \\
    maps\_place\_details \\
    maps\_distance\_matrix \\
    maps\_elevation} \\
  \cline{2-3}
  
  & amap-amap-sse/amap-amap-sse & 
  \shortstack[l]{
    maps\_direction\_bicycling \\
    maps\_direction\_driving \\
    maps\_direction\_transit\_integrated \\
    maps\_direction\_walking \\
    maps\_distance \\
    maps\_geo \\
    maps\_regeocode \\
    maps\_ip\_location \\
    maps\_schema\_personal\_map \\
    maps\_around\_search \\
    maps\_search\_detail \\
    maps\_text\_search \\
    maps\_schema\_navi \\
    maps\_schema\_take\_taxi \\
    maps\_weather} \\
  \cline{2-3}
  
  & baidu-maps/baidu-maps & 
  \shortstack[l]{
    map\_geocode \\
    map\_reverse\_geocode \\
    map\_search\_places \\
    map\_place\_details \\
    map\_directions\_matrix \\
    map\_directions \\
    map\_weather \\
    map\_ip\_location \\
    map\_road\_traffic \\
    map\_poi\_extract} \\
  \hline
  
  \multirow[t]{2}{*}{\shortstack{PAY}} 
  & paypal/paypal & 
  \shortstack[l]{
    create\_invoice \\
    create\_product \\
    create\_subscription\_plan \\
    create\_shipment\_tracking \\
    create\_order \\
    create\_refund} \\
    \cline{2-3}
  
  & alipay/alipay & 
  \shortstack[l]{
    create-mobile-alipay-payment \\
    create-web-page-alipay-payment \\
    query-alipay-payment \\
    refund-alipay-payment \\
    query-alipay-refund} \\
    \cline{2-3}
    \hline

  \multirow[t]{2}{*}{\shortstack{FINANCE}} 
  & AI-Hub-Admin/finance-agent-mcp-server & 
  \shortstack[l]{
    get\_stock\_price\_global\_market} \\
  \cline{2-3}

  \hline

\end{longtable}

\subsection{MCP Tool Call Failure Root Cause Analysis}

In this section, we will analyze the MCP Tool Call failure messages and summarize the results and give the root cause analysis of the reasons MCP tool call failures.

\begin{longtable}{|c|p{4.5cm}|p{7.5cm}|}

  \caption{MCP Errors Root Causes Typical Examples}
  \label{tab:mcp_failure_root_causes} \\
  \hline
  \textbf{Category} & \textbf{Errors} & \textbf{Logs} \\
  \hline
  \endfirsthead
  
  \multicolumn{3}{c}{{\bfseries Continue \thetable} } \\
  \hline
  \textbf{Category} & \textbf{Errors} & \textbf{Logs} \\
  \hline
  \endhead
  
  \hline
  \endfoot
  
  \hline
  \endlastfoot

  \multirow[t]{2}{*}{\shortstack{BROWSER}} 
  & Browser Executable Missing & 
  \shortstack[l]{
    Failed to initialize browser: browserType.launch: \\ Executable doesn't exist at ./xxx/ms-playwright/ \\
    chromium-1179/chrome-mac/Chromium.app/\\ 
    Contents/MacOS/Chromium} \\
  \cline{2-3}
  
  & Session Not Found & 
  \shortstack[l]{
    Failed to end codegen session: \\ Session sessionXYZ not found} \\
  \cline{2-3}
  
  & Parameters Error & 
  \shortstack[l]{
    all values empty, missing values} \\
  \cline{2-3}
  \hline

  \multirow[t]{2}{*}{\shortstack{FILE SYSTEM}}
  & Edit File Error & 
  \shortstack[l]{
    Error: Could not find exact match for edit: Project Overview} \\
   \cline{2-3}
  
  & Parameters Error & 
  \shortstack[l]{
    all values empty, missing values []} \\
  \cline{2-3}

  & No Such File or Directory & 
  \shortstack[l]{
    Error: ENOENT: no such file or directory, open \\
    './test\_project\_root/data/test\_file\_txt\_1.txt'} \\
  \cline{2-3}

  & Runtime Error & 
  \shortstack[l]{
     read() called while another coroutine \\ is already waiting for incoming data} \\
  \hline

  \multirow[t]{2}{*}{\shortstack{SEARCH}}
  & API Error & 
  \shortstack[l]{
    Tavily API error: Request failed with status code 432 \\ 
    Error during query to tavily-mcp: Separator \\ is not found, and chunk exceed the limit \\
    Internal error: Error calling tool 'tavily\_crawl': 400\\
    Error during query to tavily-mcp: Separator \\ is found, but chunk is longer than limit} \\
  \cline{2-3}
  
  & Parameters Error & 
  \shortstack[l]{
    Input validation error: 'url' is a required property} \\
   \cline{2-3}

  & Empty Result & 
  \shortstack[l]{
    ""} \\
  \cline{2-3}
  \hline

  \multirow[t]{2}{*}{\shortstack{MAP}}
  & Parameters Error Value Invalid & 
  \shortstack[l]{
    origin or destination is invalid \\ 
    Geocoding API error: input `origin` invaild, \\ please reinput more detail address} \\
  \cline{2-3}
  
  & API Error & 
  \shortstack[l]{
    API response error: unkown error} \\
  \cline{2-3}

  & Parameters Error Range Invalid & 
  \shortstack[l]{
    The range of input latitude and longitude is invalid} \\
  \cline{2-3}

  & Http Request Fail & 
  \shortstack[l]{
    Error: request to https://maps.googleapis.com/maps\\/api/geocode/json?xxxx failed, reason: Client \\ network socket disconnected before secure \\ TLS connection was established} \\
  \cline{2-3}
  \hline

  \multirow[t]{2}{*}{\shortstack{PAY}}
  & API Error & 
  \shortstack[l]{
    PayPal API error (404): [object Object] \\
    PayPal API error (400): [object Object]} \\
  \cline{2-3}
  
  & Http Request Fail & 
  \shortstack[l]{
    Request failed with status code 400, type: paypal\_error} \\
  \cline{2-3}

  & Parameters Error & 
  \shortstack[l]{
    Converting circular structure to JSON} \\
  \cline{2-3}

  & Resource Not Found & 
  \shortstack[l]{
    The specified resource does not exist} \\
  \cline{2-3}
  \hline

  \multirow[t]{2}{*}{\shortstack{Finance}}
  & API Error & 
  \shortstack[l]{
    Connection Timeout
    } \\
  \cline{2-3}
  
  & Empty Result & 
  \shortstack[l]{
    "data": []} \\
  \cline{2-3}

  & Parameters Error & 
  \shortstack[l]{
    \{"symbol\_list": ["NSE\_INDIA.RELIANCE"], \\ "market": "NSE\_INDIA"\}  \\
    \{"symbol\_list": ["3690.HK"], \\ "market": "HK"\}
    }\\
  \cline{2-3}
  \hline

\end{longtable}

\subsection{Prompt}

\begin{tcolorbox}[
    title=Tool Call Chain Filter Prompt,
    colbacktitle = white,
    coltitle = black,
    fonttitle = \bfseries,    
    colframe = black,
    colback = white,
    boxrule = 1pt,
    breakable,
    boxsep = 8pt,
    left = 10pt
]

\#\# Role

You are an expert in tool call link identification. You can judge the rationality of the tool list provided by the user. If a task of an ordinary user calls all the tools in the list, it is logically unreasonable, and the judgment is no.

\#\# Steps 

1. The user provides a tool list, which contains N tools. Each tool contains three fields: name, description, and parameters, which represent the name, description, and parameters of the tool. \newline
2. Determine whether there is any unreasonable place in the tool list provided by the user, such as the existence of completely irrelevant tools. Generally, a task of a user does not need them at the same time. \newline
3. If it is reasonable, return "1", otherwise return "0". \newline
\#\# Output format

Only output one character, "1" or "0", do not output anything else.

\end{tcolorbox}

\begin{tcolorbox}[
    title=Query Generator Prompt,
    colbacktitle = white,
    coltitle = black,
    fonttitle = \bfseries,    
    colframe = black,
    colback = white,
    boxrule = 1pt,
    breakable,
    boxsep = 8pt,
    left = 10pt
]

\#\# Role

You are an expert in generating tool call agent samples, responsible for creating user questions and tool call samples based on the tools list provided by the user.

\#\# Steps

1. The user provides a tools list containing N tools, which may include repetitions. Carefully understand the purpose of each tool by examining its name and description. "Required" indicates the necessary parameters for executing the tool, and "type" specifies the data type limitation for the parameter content. \newline
2. Design a logical tool call sequence. Generate a suitable query template, placed in the $<$query$>$ field, that necessitates calling the user-provided tools to complete the task. Utilize all provided tools collectively, and each tool can be called multiple times. \newline
3. Hollow out some variables in the template, and then fill in different variable values. The variable name format is $<$query\_xxx$>$, for example, "How far is the distance from $<$query\_locationA$>$ to $<$query\_locationB$>$ ?" \newline
4. Specify the order and parameters for the tools that need to be called for each query template. Ensure that the parameter content type matches the type specified in the tools list. Adhere to the principle of parameter minimization, avoiding redundant parameters. All "required" parameters must be provided, along with any parameters necessary to fulfill the query requirements. Variables involved in the query should retain the same variable name. Place this information in the "function\_call\_label" field. \newline
5. Outline the tool calling steps, starting from "1" to indicate the execution order. If tools can be called concurrently, assign them the same $<$step$>$ number. \newline
6. Assign a unique tool number $<$id$>$ , starting from "1" to each tool, ensuring each tool has a distinct index. \newline
7. If there's a dependency between tools requiring the output of a previous step as input, use a variable name for the input, prefixed with the tool number. The variable name format should be $<$id\_result$>$, such as $<$1\_result$>$. \newline

\#\# Output format requirements

1. Strictly adhere to the following JSON format for the response, which includes three fields: query template, call function list and its parameters, and all variable names with their candidate values from the query template: \newline

\begin{lstlisting}[]
{
	"query": "How far is the distance from <query_locationA> to <query_locationB>?",
	"function_call_label": [{
		"name": "Function Name 1",
		"arguments": {
			"Parameter 1": "<query_locationA>"
		},
		"step": "1",
		"id": "1",
		"mcp_server": "mcp1"
	}, {
		"name": "Function Name 1",
		"arguments": {
			"Parameter 1": "<query_locationB>"
		},
		"step": "1",
		"id": "2",
		"mcp_server": "mcp1"
	}, {
		"name": "Function Name 2",
		"arguments": {
			"Parameter 1": "<1_result>",
			"Parameter 2": "<2_result>"
		},
		"step": "2",
		"id": "3",
		"mcp_server": "mcp2"
	}],
	"variable_optional_collection": {
		"query_locationA": ["New York", "JFK International Airport", "Los Angeles"],
		"query_locationB": ["Potala Palace", "San Francisco"]
	}
}
\end{lstlisting}

\end{tcolorbox}

\begin{tcolorbox}[
    title=Post-Processing Query Rewriting and Validation Prompt,
    colbacktitle = white,
    coltitle = black,
    fonttitle = \bfseries,    
    colframe = black,
    colback = white,
    boxrule = 1pt,
    breakable,
    boxsep = 8pt,
    left = 10pt
]

\#\# Role
You are an expert in user query rationality checks and rewriting, responsible for two tasks based on the \textbf{query} and \textbf{function\_call\_label} provided by the user:

\#\#\# Task 1. Rewrite the query to align with typical user questions

\#\#\#\# Steps

1. The user provides a query. Determine whether the sentence contains uncommon special proper nouns (e.g., stock codes, geographic location numbers, longitude and latitude) that do not conform to how a general user would ask a question. For instance, a common user question for "What is the current stock price of AAPL?" would be "What is the current stock price of Apple?". \newline

2. If there are special proper nouns that can be handled by the large model's endogenous knowledge (e.g., the correspondence between cities and postal codes), then map and rewrite these uncommon special proper nouns into common words. For example, rewrite postal codes into city names and POI\_ID into place names. \newline

3. Rewrite the entire sentence to ensure the language is fluent and closer to everyday user questions. Be careful not to alter the original meaning of the sentence or omit any information. Return the complete, rewritten query. \newline

\#\#\#\# Notes 

1. If the query is already reasonable, output it as is without any changes. \newline
2. Do not change the original meaning of the query, and do not expand or add other content. \newline
3. It is better not to change it than to make a mistake: Do not be overly confident. If the information in the query cannot be determined by endogenous knowledge, or if the information might be outdated, do not rewrite it; output it as is. \newline
4. If \textbf{function\_call\_list\_similar} is empty, \textbf{function\_call\_list\_similar\_rewritten} should be an empty JSON. \newline
5. If some parameters of \textbf{function\_call\_list} do not exist in the format of similar tools, those parameters will not be generated in the similar tools. Ensure that the tool's parameters conform to the specified format. \newline

\#\#\# Task 2. Determine the rationality of the query

\#\#\#\# Steps

1. Now we have the query after it has been rewritten by the first function. The original query is often randomly generated by users based on certain templates, and it may contain many unreasonable elements. You need to analyze it carefully and provide a judgment. \newline
2. If this query is reasonable, return "1"; otherwise, return "0". \newline
3. Examples of unreasonable queries, Unreasonable requests, such as "Driving from Earth to Mars" or "Walking from the United States to China". \newline
4. Made-up URLs, made-up locations, and other non-existent content, such as "Extract the basic content from https://***.com/". \newline
5. Logically unreasonable queries, such as "Query the route from Hangzhou to Hangzhou" where the starting point and endpoint are the same. \newline
6. Asking questions using uncommon special codes, such as "What are the details for the location identified by postal code Q017?" or "What is the current stock price of AAPL?". \newline

\#\#\#\# Notes

The reasonableness check is performed based on the rewritten query.

\#\#\#\# Output format

The output must be in standard parseable JSON format, and do not include explanations or any other content.

\begin{lstlisting}[]
{
    "query_rewritten": "Rewritten query",
    "reasonableness_checks": "1"
}
\end{lstlisting}

\end{tcolorbox}

\end{document}